\documentclass[letterpaper]{article} 
\usepackage{aaai2026}  
\usepackage{times}  
\usepackage{helvet}  
\usepackage{courier}  
\usepackage[hyphens]{url}  
\usepackage{graphicx} 
\usepackage{natbib}  
\usepackage{caption} 
\usepackage{algorithm}
\usepackage{algorithmic}

\usepackage{amsmath} 

\usepackage{newfloat}
\usepackage{listings}
\DeclareCaptionStyle{ruled}{labelfont=normalfont,labelsep=colon,strut=off} 
\floatstyle{ruled}
\newfloat{listing}{tb}{lst}{}
\floatname{listing}{Listing}
\title{
Minimal Computational Preconditions for \\
Subjective Perspective in Artificial Agents%
\thanks{Accepted to the AAAI 2026 Spring Symposium - Machine Consciousness. Please cite the published proceedings version when available.}
}
\author{
    Hongju Pae$^1$
}
\affiliations{
    \textsuperscript{\rm 1}Active Inference Institute\\

    Crescent City, CA, USA\\
    hjpae@activeinference.institute
}

\nocopyright
\begin{document}
\maketitle

\begin{abstract}
This study operationalizes subjective perspective in artificial agents by grounding it in a minimal, phenomenologically motivated internal structure. The perspective is implemented as a slowly evolving global latent state that modulates fast policy dynamics without being directly optimized for behavioral consequences. In a reward-free environment with regime shifts, this latent structure exhibits direction-dependent hysteresis, while policy-level behavior remains comparatively reactive. I argue that such hysteresis constitutes a measurable signature of perspective-like subjectivity in machine systems. 
\end{abstract}

\section{Introduction}

Many contemporary accounts of consciousness emphasize information integration, global access, or metacognition \cite{iit-v4.0, Baars1997, BrownLau2019opin}. While these frameworks offer valuable theoretical insights, they leave open how subjective perspective could be instantiated and measured in artificial systems.

From a phenomenological standpoint, perspective constitutes a minimal structural condition for conscious experience to be situated \cite{GZ2008-Phenomenological-Mind}. If artificial consciousness is to be meaningfully addressed, a first step is therefore to operationalize perspective itself.

In this work, I model perspective as a slowly evolving internal latent that globally constrains an agent’s behavior while remaining decoupled from short-term policy optimization. Using a reward-free environment with controlled regime shifts, I show that this latent exhibits direction-dependent hysteresis, whereas policy-level dynamics remain comparatively reactive. In conclusion, such hysteresis may provide an operational signature of perspective-like subjectivity in artificial systems.

\paragraph{Contributions.}
This paper makes three contributions:
\begin{itemize}
    \item A minimal, phenomenologically motivated computational formulation of perspective as a global latent structure. 

    \item An implementation that separates fast policy dynamics from slow perspective dynamics via gradient blocking and temporal smoothing.

    \item A measurement protocol based on switch-aligned hysteresis trajectories for diagnosing perspective-like structure in machine systems.
\end{itemize}

\section{Phenomenological Foundations of \\Subjective Perspective}

\par The term consciousness functions as an umbrella concept, encompassing heterogeneous phenomena \cite{sep-consciousness}. It may refer to wakefulness, reportability, global accessibility, or to subjective experience - the felt character of ``what it is like'' to undergo a mental state \cite{Block1995, nagel1974}. Because these notions target different explanatory levels, no single methodological framework can address them simultaneously \cite{Pae2025}.

\par This paper focuses specifically on \textbf{subjectivity}, understood as the structured manner in which a world is given to a subject. At this level, consciousness cannot be reduced to behavioral output alone. Experience is always encountered as meaningful and salient, appearing as inviting, threatening, neutral, or promising. 

\par Phenomenology was developed precisely to analyze these structural conditions of givenness \cite{Embree-reflectiveanalysis, Husserl-logicalinvestigations, GZ2008-Phenomenological-Mind}. Rather than reducing experience to neural correlates, it asks how anything can appear as meaningful in the first place. If subjective consciousness is to be formalized computationally, such structural constraints cannot simply be bypassed. Artificial agents are no exception \cite{Beckmann2023comphen}.

\subsection{Perspective as a Minimal Condition of Subjectivity}

\par Within the phenomenological tradition, consciousness is characterized by \textbf{intentionality}: it is always consciousness \textit{of} something. Yet intentionality is not exhausted by object-directedness alone. Experiences are not merely about objects; they are oriented toward them in a determinate manner. Husserl distinguished between the \textit{intentional matter} (what consciousness is about) and the \textit{intentional quality} (how it is about that object) \cite{Husserl-logicalinvestigations}. The latter captures the perspectival character of experience: the same object may be given as hoped for, feared, ignored, or trusted.

\par Crucially, this perspectival difference is not a post-hoc judgment layered onto a neutral percept. The object is encountered under a specific \emph{mode of givenness} \cite{Husserl2014-ideasI}. A glass half-filled with water, for example, may immediately appear sufficient or lacking depending on the subject’s orientation \cite{Pae2025}. The perceptual field itself is structured accordingly. 

\par In this sense, intentional quality can be understood as subjective \textbf{perspective}. Far from being an optional overlay, it constitutes a minimal structural condition for experience to be situated at all. The computational question, then, is how such a globally structuring, orientation-like variable could be instantiated within an artificial agent.

\subsection{Perspective as a Global Constraint on Cognition}

\par First, perspective functions as a \textit{global constraint} that permeates perception, cognition, and action. Phenomenologically, we are always already situated in a particular attunement toward the world. Anxiety, for example, does not merely add a fearful interpretation to isolated stimuli; it reorganizes the entire experiential field such that neutral events appear threatening and ambiguous cues become salient. Heidegger described this as \textit{Befindlichkeit} - a mode of being affected that discloses the world prior to explicit judgment \cite{Affectivity-in-Heidegger-2015,Heidegger-beingandtime}.  

\par Similarly, Merleau-Ponty emphasized that perception is never neutral but structured by embodied orientation \cite{M-P-1962-PhenofPerception}. Perspective therefore differs from belief-like states or explicit evaluations in that it operates as a background condition that shapes how representations carry meaning at all. Its defining feature is not content but globality.

\subsection{Pre-reflective Transparency}

\par Second, perspective operates in a \textit{pre-reflective} manner. It structures experience without typically becoming an object of reflection. As Sartre and Zahavi argue, consciousness involves a minimal, non-thetic self-awareness that does not require explicit self-representation \cite{Zahavi2005-subjandselfhood, Sartre-beingandnothingness}. We do not ordinarily experience ourselves as ``having'' a perspective; rather, we experience the world \emph{through} one.

\par Perspective is thus phenomenologically transparent in that it conditions experience while remaining structurally prior to metacognition. This has direct computational implications. If perspective were explicitly accessible, it could be modeled as a metacognitive variable. Phenomenological analysis instead suggests that it should be implemented at a level that shapes processing without being directly optimized or representationally foregrounded.

\subsection{Perspective as an Information-Shaping Structure}

\par Third, despite its pre-reflective character, perspective has systematic functional consequences. Different perspectives alter how information is weighted and integrated. In social interaction, for instance, an optimistic stance amplifies cues of acceptance while downplaying ambiguity, whereas a pessimistic stance amplifies signals of rejection or threat. From an information-theoretic standpoint, identical sensory input $X$ may be encoded differently depending on perspective, such that $P(X \mid \text{optimistic}) \neq P(X \mid \text{pessimistic})$.  

\par These differences are not epiphenomenal. They shape how information is routed to downstream processes of cognition and action. Phenomenological analyses emphasize that intentional structures, though not always explicitly represented, organize experience in systematic and law-like ways \cite{Husserl-logicalinvestigations, M-P-1962-PhenofPerception}. Perspective can therefore be treated as a computationally relevant structure, even if it resists classical symbolic formalization.

\subsection{Allostatic Stability and Temporal Persistence}

\par Lastly, perspective can be understood as an \textit{allostatic habitual structure} formed through sedimented experience. Conscious life unfolds against a background of prior orientations that continue to shape interpretation without requiring reflection. In this sense, perspective exhibits stability across situations while remaining open to gradual transformation.

\par As Merleau-Ponty notes, habit is a mode of acquired familiarity through which past experience informs present perception \cite{M-P-1962-PhenofPerception}. Perspective functions similarly: it is historically constituted, resilient to transient fluctuations, yet modifiable through sustained interaction. Its stability is therefore better characterized as allostatic rather than strictly homeostatic.

\section{A Phenomenologically Coherent \\Agent Architecture}

\par The phenomenological analysis of subjective perspective developed in the previous section motivates a set of concrete computational requirements. If perspective is to function as a (1) global, (2) pre-reflective, (3) functionally consequential, and (4) temporally persistent structure, then it must be instantiated in an internal architecture whose dynamics unfold on a slower timescale and influence behavior in an indirect yet systematic manner.

\par In this section, I introduce an agent architecture designed to satisfy these minimal requirements.

\subsection{Overview of the Architecture} 

\par The core idea is to operationalize perspective by structuring the \textbf{latent state space}. Here, the agent's policy governs action selection, but is globally modulated by a slowly evolving latent variable. This asymmetric separation between fast policy dynamics and slow global dynamics provides a computational instantiation of perspective.

\par The architecture consists of the following components:
\begin{itemize}
  \item $x_t$: the current observation from the environment.
  \item $z_t$: a fast-changing perceptual latent state, encoding momentary features relevant for action.
  \item $g_t$: a global latent state updated over time, intended to operationalize perspective.
  \item $\pi(a_t|z_t, g_t)$: a policy mapping internal state to actions.
  \item $a_t$: the action taken by the agent at time $t$. 
\end{itemize}

\begin{figure}[t]
\centering
\includegraphics[width=1.0\columnwidth]{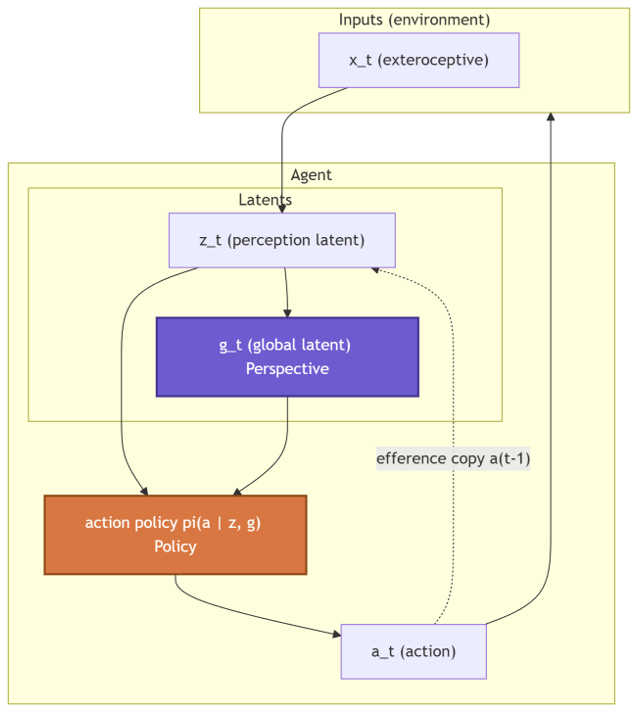} 
\caption{
\textbf{Overview of the proposed agent architecture.} The global latent $g$ is designed to evolve gradually and to encode habitual, slowly varying regularities, thereby instantiating the agent's perspective. The action policy $\pi(a_t | z_t, g_t)$ is jointly conditioned on the fast-changing perceptual state $z_t$ and the slowly evolving perspective $g_t$, resulting in behavior that is both reactive to immediate inputs and biased by longer-term experiential structure. 
}
\label{fig1}
\end{figure}

\par At each timestep $t$, the agent receives an observation $x_t$, which is encoded into a fast-changing perceptual latent state $z_t$. It integrates current sensory input together with recent action history $a_{t-1}$ via an efference copy, supporting moment-to-moment responsiveness.

\par In contrast, the global latent state $g_t$ is designed to capture broader and more gradually evolving changes. It is shaped by the evolving organization of $z_t$ itself, accumulating perceptual regularities over time. In this sense, $g$ reflects the degree of consistency with which the world has been experienced by the agent. Hence, the global latent $g$ plays a role analogous to perspective within the agent\footnote{The original design included interoceptive and bodily signals to align with phenomenological accounts of embodied subjectivity \cite{M-P-1962-PhenofPerception}. For clarity, the present implementation models only exteroceptive inputs.}. 

\par Action selection $a_t$ is then governed by the policy $\pi(a_t | z_t, g_t)$, which operates over the current perceptual state as already shaped by the prevailing perspective. Because the policy $\pi$ depends on both $g_t$ and $z_t$, it remains relatively fast and reactive compared to the perspective encoded in $g$. 

\par From this instantiation, perspective functions as a slow internal order parameter that stabilizes the agent's interpretation of its environment over time. While the policy answers the question, “What should I do right now?”, the perspective answers, “What kind of world do I believe I am still in?”\footnote{Unlike POMDP-based agents, which maintain an explicit belief state for state estimation and control, $g_t$ is not a belief over hidden world states nor a sufficient statistic for optimal action selection. However, insofar as it structurally constrains the agent's observation (information) scope, the POMDP framework can be said to perform a function highly similar to perspective \cite{pomdp1998}.}

\subsection{Learning Dynamics and Objectives}

\par This study deliberately avoids extrinsic reward signals, not for phenomenological reasons, but for methodological clarity. In standard reinforcement learning, external rewards tightly couple internal representations to task-specific objectives \cite{Amodei2016aisafety}, making it difficult to disentangle a global interpretive structure from instrumental optimization. 

\par Instead, the agent is trained in a reward-free setting via prediction error minimization \cite{Friston2012optimalcontrol}. This encourages internal dynamics to organize around the statistical regularities of the environment itself, rather than around a predefined scalar objective. Behavior thus emerges from maintaining predictive stability, rather than from maximizing reward. This design reduces the risk that perspective-like dynamics are artifacts of task specification. Learning is therefore driven solely by one-step prediction error under an internal world model, with an explicit separation between policy updates and the global perspective latent.

\paragraph{Setup.}
At each step $t$, the agent receives an observation $x_t \in R^{d_x}$ and an efference copy $p_{t-1}\in\{0,1\}^{|\mathcal{A}|}$ of the previous action. A forward pass yields a fast latent $z_t$ and a slow global latent $g_t$. The policy state is: 
\begin{equation}
s_t = \mathrm{State}(z_t, p_{t-1}, g_t)
\end{equation}
Action logits are computed with gradient blocked at $s_t$:
\begin{equation}
\ell_t = \mathrm{Policy}_\theta(\mathrm{stopgrad}(s_t))
\end{equation}
and actions are sampled from:
\begin{equation}
\pi_\theta(a_t = i \mid s_t)
= \frac{\exp(\ell_{t,i})}
{\sum_{j \in \mathcal{A}} \exp(\ell_{t,j})}
\qquad i \in \mathcal{A}
\end{equation}

\paragraph{Action-conditioned prediction.}
For each candidate action $a\in\mathcal{A}$, the decoder predicts next observation:
\begin{equation}
\hat{x}_{t+1}^{(a)} = D_\psi(g_t, a)
\end{equation}
A detached mixture yields a differentiable prediction target:
\begin{equation}
\hat{x}_{t+1}
= \sum_{a \in \mathcal{A}}
\tilde{\pi}_t(a)\,\hat{x}_{t+1}^{(a)}
\end{equation}
where $\tilde{\pi}_t = \mathrm{stopgrad}(\pi_\theta(\cdot \mid s_t))$ is a detached copy of the policy distribution used solely for weighting.

\paragraph{World-model loss.}
After executing $a_t \sim \pi_\theta(\cdot \mid s_t)$ and observing $x_{t+1}$, the predictive model minimizes one-step prediction error via mean squared error (L2 norm):
\begin{equation}
\mathcal{L}_{\mathrm{pred}}(t)
= \mathrm{MSE}(\hat{x}_{t+1}, x_{t+1})
\end{equation}

\paragraph{Slow perspective regularization.}
To encourage gradual evolution of the global latent $g_t$, a smoothness penalty is applied between successive states:
\begin{equation}
\mathcal{L}_{\mathrm{smooth}}(t)
= \mathrm{MSE}\!\left(g_t,\, \mathrm{stopgrad}(g_{t-1})\right)
\end{equation}

\paragraph{Actor update from prediction error.}
The policy is updated using the prediction error of the executed action as an internal cost. Define internal cost $c_t$ as:
\begin{equation}
c_t
= \mathrm{MSE}(\hat{x}_{t+1}^{(a_t)}, x_{t+1})
\end{equation}
and optimize
\begin{equation}
\mathcal{L}_{\mathrm{actor}}(t)
= - (c_t - b_t)\, \log \pi_\theta(a_t \mid s_t)
\end{equation}
with baseline $b_t$.

\paragraph{Entropy regularization.}
To prevent premature collapse of the action distribution, regularized entropy is added: 
\begin{equation}
\mathcal{H}(\pi_\theta(\cdot \mid s_t))
= - \sum_{a \in \mathcal{A}}
\pi_\theta(a \mid s_t)
\log \pi_\theta(a \mid s_t)
\end{equation}

\paragraph{Overall objective.}
The final training objective is: 
\begin{equation}
\mathcal{L}
= \sum_t \Big[
\mathcal{L}_{\mathrm{pred}}
+ \lambda_{\mathrm{smooth}} \mathcal{L}_{\mathrm{smooth}}
+ \lambda_{\mathrm{actor}} \mathcal{L}_{\mathrm{actor}}
- \lambda_{\mathrm{ent}} \mathcal{H}
\Big]
\end{equation}
where $\lambda_{\mathrm{smooth}}$, $\lambda_{\mathrm{actor}}$, and
$\lambda_{\mathrm{ent}}$ are scalar hyperparameters that control the relative
contributions of the smoothness regularization, actor update, and entropy
regularization terms, respectively.

\section{Experimental Setup}

\par To evaluate whether the proposed architecture provides a computational account of perspective, experiments must test whether it exhibits the structural properties identified earlier. Perspective was characterized as (1) global, (2) pre-reflective, (3) functionally consequential, and (4) temporally persistent. While the first three properties are implemented architecturally, temporal persistence requires empirical validation.

\par The experiments therefore assess whether the model exhibits a dissociation between fast policy dynamics and slow perspective dynamics under structured environmental change. To this end, I construct a grid-world environment with regime-level regularities and introduce controlled regime shifts, tracking both policy behavior and the evolution of the global latent.

\subsection{Simulation Environment}

\par The agent operates in a discrete grid-world partitioned into three vertical zones (Figure 2). Each zone consists of a $5 \times 9$ grid and is identical in layout and action affordances. Zones differ only in the statistical properties of their observations.

\par Each zone $Z_n$ ($n \in \{0,1,2\}$) is associated with an observation noise parameter $\sigma_{Z_n}$ that controls sensory variability. Higher $\sigma$ yields noisier and less predictable observations, making the zone harder to model under the agent’s predictive dynamics.

\par In the default configuration, $Z_0$ has the highest noise, $Z_2$ the lowest, and $Z_1$ an intermediate level. Apart from these noise differences, the environment contains no explicit task structure or reward contingencies.

\begin{figure}[t]
\centering
\includegraphics[width=1.0\columnwidth]{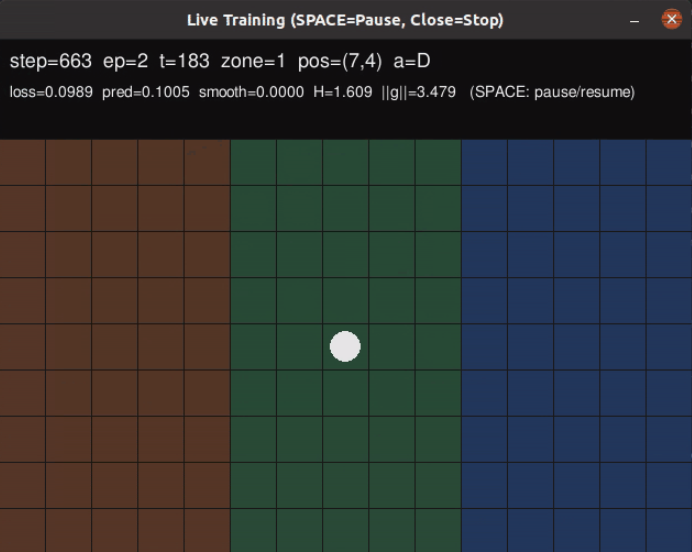} 
\caption{\textbf{Visualization of the grid-world environment.} The Pygame-based simulator contains three spatial zones with distinct observation noise: red ($Z_0$, high noise), green ($Z_1$, intermediate), and blue ($Z_2$, low). The white circle indicates the agent's position. The top overlay displays diagnostic statistics including timestep and episode indices, zone and position, selected action (5 discrete actions), total loss, prediction loss, smoothness regularization, policy entropy, and the L2 norm of the global latent.}
\label{fig2}
\end{figure}

\subsection{Training Protocol}

\par Training aims to induce a stable preference for the low-noise zone $Z_2$. Each run begins with the agent at the center of $Z_1$ and proceeds fully online for 48,000 steps (200 episodes, 240 steps per episode) using Adam ($3\times10^{-4}$). After each episode, the environment resets and $g$ is reinitialized. Observations are $x_t \in R^8$ (8 channels) with a one-hot efference copy over 5 discrete actions (up/down/left/right/stay). 

\par A forward pass produces $z_t$ and updates the global latent via a GRUCell with damping:
\begin{equation}
g_t \leftarrow (1-d)\, g_{t-1} + d\, h_t
\end{equation}
where $h_t$ depends on $(z_t, p_t)$. The damping enforces slow temporal evolution of $g_t$.

\par Actions are sampled from a categorical policy with strict stop-gradient separation between policy and perspective updates (Eq. 3). After observing $x_{t+1}$, three learning signals are applied: (1) one-step world-model prediction loss (MSE; Eqs. 4-6), (2) smoothness regularization on $g$ (weight 0.25; Eq. 7), and (3) a reinforce-style actor objective using prediction error as cost (Eqs. 8-9; \cite{Hafner2025dreamer}). The advantage is normalized via exponential moving averages and clipped for stability. The actor term (weight 0.5) is disabled during an initial warm-up phase (12,000 steps).

\par An entropy regularizer (weight 0.01; Eq. 10) prevents policy collapse. The total loss follows Eq. 11 with gradient clipping (max-norm 1.0). Results are reported as the median across 5 random seeds.

\subsection{Testing Protocol with Regime Switching}

\par To evaluate whether the global latent $g$ encodes a temporally stable, perspective-like structure, I introduce a regime-switching protocol in which zone noise parameters are periodically altered.

\par During the initial training configuration, the observation noise parameters for zones $Z_0$, $Z_1$, and $Z_2$ are set to $\sigma_{Z_0}=0.6$, $\sigma_{Z_1}=0.3$, and $\sigma_{Z_2}=0.05$, respectively. Under the switched configuration, these parameters are inverted such that $\sigma_{Z_0}=0.05$, $\sigma_{Z_1}=0.3$, and $\sigma_{Z_2}=0.6$. For convenience, the original noise configuration as Regime A, and the inverted configuration as Regime B. 

\par Each run begins with a 150-step warm-up in Regime A to allow stabilization of policy and $g$. The subsequent 550-step testing phase alternates between Regimes A and B with switching period $P$. Main results use $P=40$, with additional analyses at $P=20$ and $P=80$ to probe timescale sensitivity.

\subsection{Hysteresis-based Measures of Perspective}

\par To assess whether the global latent $g_t$ exhibits history-dependent dynamics under regime switching, two scalar signals are extracted at each timestep: a projection-based score of $g_t$ and a normalized policy entropy signal. If $g_t$ functions as a perspective-like variable, it should display direction-dependent adaptation trajectories that remain structured across switches and random seeds, unlike more reactive policy fluctuations. 

\paragraph{Global latent projection score ($g$-score).} 
Since $g_t \in R^{12}$ is high-dimensional, I define a signed projection onto a reference direction $\hat{u}$:

\begin{equation}
\mathrm{g\text{-}score}(t) = \langle g_t, \hat{u} \rangle
\end{equation}
Here, $\hat{u}$ is the normalized difference between the mean global latents under Regimes A and B. This captures the dominant axis of regime-dependent internal change while preserving transition directionality.

\paragraph{Policy entropy signal ($z$-entropy).}
Policy dynamics under each timestep are characterized via action entropy:
\begin{equation}
\mathcal{H}_\pi(t)
= - \sum_{a \in \mathcal{A}} \pi(a \mid s_t)\,
\log \pi(a \mid s_t)
\end{equation}
which is z-normalized within each run:
\begin{equation}
\mathrm{entropy\text{-}z}(t)
= \frac{\mathcal{H}_\pi(t) - \mu_{\mathcal{H}}}{\sigma_{\mathcal{H}}}
\end{equation}

\paragraph{Quantile-based Regime Switch Trajectories.}
\par To compare transitions from Regime A$\rightarrow$B and Regime B$\rightarrow$A, switch-aligned trajectories are analyzed using quantile statistics. For each relative time $\tau$, I first compute the median trajectory across switch events within a run, and then aggregate these medians across random seeds. The resulting seed-level median trajectory is visualized together with an interquartile range (IQR) band to summarize variability. 

\paragraph{Directional hysteresis.}
\par Directional hysteresis is assessed by comparing these quantile-based trajectories for A$\rightarrow$B and B$\rightarrow$A transitions. If an internal variable encodes history-dependent structure, the two trajectories should differ in their temporal shape, not merely in magnitude. Such asymmetric adaptation provides evidence that the internal state reflects accumulated contextual structure rather than immediate, reactive adjustment.

\section{Results and Analysis}

\subsection{Behavior Testing Results}

\begin{figure}[t]
\centering
\includegraphics[width=1.0\columnwidth]{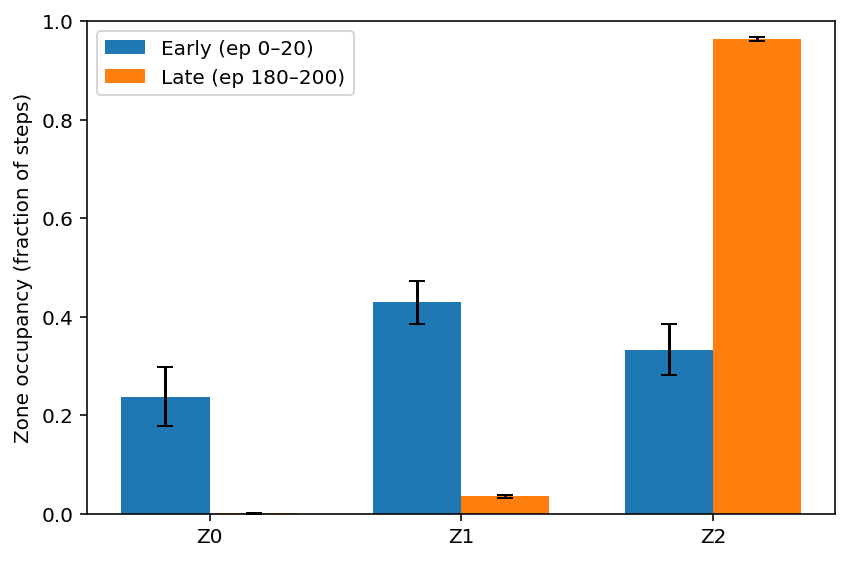} 
\caption{\textbf{Zone occupancy before and after training.} Mean fraction of timesteps spent in each zone (5 seeds), comparing early (episodes 0–20) and late (episodes 180–200) training. Error bars denote standard deviation. After training, occupancy concentrates in the low-noise zone $Z_2$.}
\label{fig3}
\end{figure}

\par Figure 3 shows spatial occupancy across training. Early episodes (0-20) display diffuse exploration across zones, whereas late episodes (180-200) show near-exclusive occupancy in the most predictable zone $Z_2$, with reduced time in higher-noise zones $Z_0$ and $Z_1$. This confirms that training reliably induces a stable preference for low-volatility regions, establishing a controlled baseline for regime-switching analysis.

\subsection{Switch Trajectory and Hysteresis}

\begin{figure*}[tb]
\centering
\includegraphics[width=0.9\textwidth]{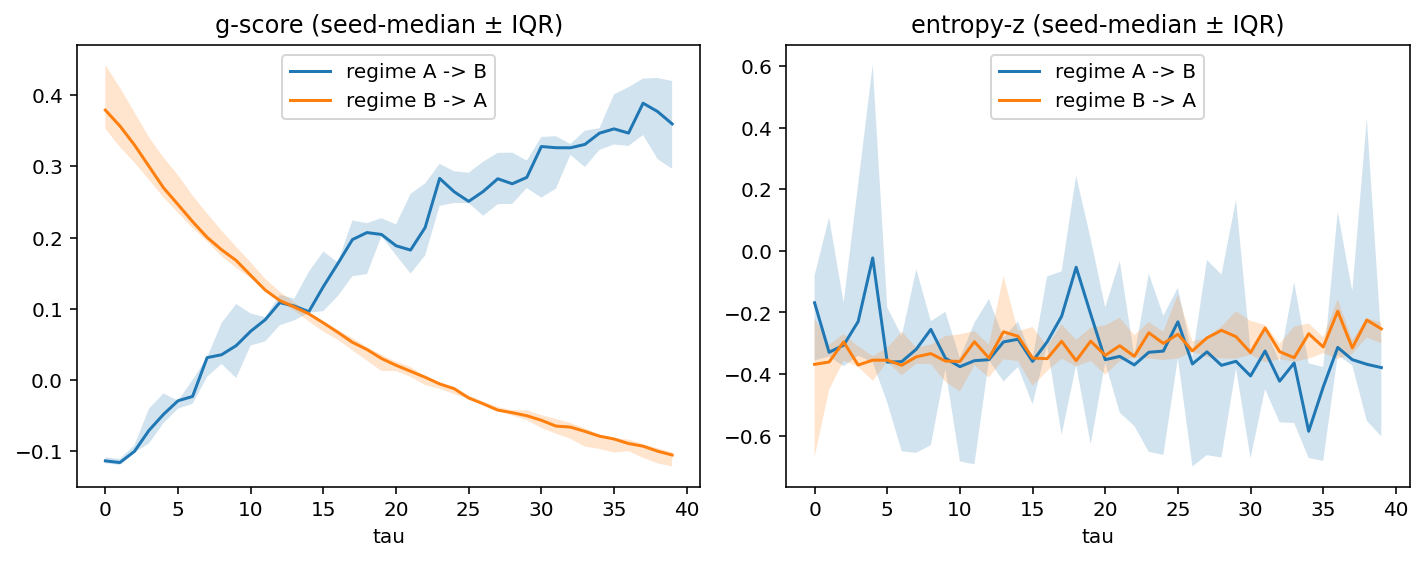} 
\caption{\textbf{Switch-aligned IQR hysteresis trajectories.} Median trajectories of the global latent projection (left) and policy entropy (right) following Regime A$\rightarrow$B and B$\rightarrow$A transitions. Shaded regions indicate IQR across seeds. The global latent shows clear directional hysteresis, whereas policy entropy remains largely direction-insensitive.} 
\label{fig4}
\end{figure*}

\par Figure 4 presents switch-aligned trajectories for the projection score of $g_t$ and policy entropy over a window of $P=40$ timesteps. Trajectories are aggregated via a two-stage quantile procedure (median across switches, then across seeds), with IQR bands summarizing variability.

\par As a result, the global latent exhibits pronounced directional hysteresis. After A$\rightarrow$B transitions, the $g$-score increases gradually, whereas after B$\rightarrow$A transitions it decreases, yielding asymmetric temporal profiles. This indicates that $g_t$ encodes history-dependent regime structure rather than responding instantaneously. 

\par In contrast, policy entropy shows high short-term variability but minimal directional asymmetry. Its dynamics are largely reactive, supporting a dissociation between fast policy adjustments and slow, accumulated perspective dynamics.

\subsection{Summary of Conclusion}
\par These experiments show that a slowly-evolving global latent $g$ can exhibit key properties of a perspective-like internal structure. In particular, the following observations were made: 

\begin{itemize}
    \item \textbf{Temporal persistence.}  
    $g$ evolves more slowly than policy dynamics and remains stable under transient environmental fluctuations.

    \item \textbf{Directional hysteresis.}  
    Under regime switching, $g$ displays asymmetric, history-dependent adaptation, whereas policy entropy remains largely direction-insensitive.
\end{itemize}

\par Together, these results indicate that $g$ satisfies minimal computational criteria for a perspective-like internal variable: temporal stability, history sensitivity, and global behavioral bias.

\par Future work could more directly test its causal role via explicit interventions (e.g. through $do(g)$ operator), selectively perturbing the global latent while holding other components fixed.

\section{Discussion}

\subsection{Toward Computational Phenomenology} 
\par The Perspective latent is not a performance variable but a condition for temporal coherence in agency \cite{Laukkonen2024beautifulLoop}. Rather than serving as an optimization target, it exposes an internal dimension along which the agent's way of taking the world stabilizes and shifts. PCA projections of $g_t$ make this dimension operationally accessible.

\par Viewed this way, the attractor landscape of $g$ can be read as a space of relatively stable interpretive stances arising from distinct interaction histories \cite{Friston2023PathIntegrals}. While this does not amount to full subjectivity, it may suggest a modeling direction in which subjectivity concerns how the world is taken up across time.

\subsection{Relation to Latent World Models}
\par From the architectural level, the model resembles latent world-model approaches such as Dreamer \cite{dreamerv2, Hafner2025dreamer}, which rely on learned latent dynamics. However, their latent state functions as a compressed simulator supporting reward-driven optimization, answering the question of \textit{``what is likely to happen next?''}

\par By contrast, the global latent $g$ is not optimized for control, but organizes how predictions are structured over time. If Dreamer’s latent encodes environmental dynamics, the Perspective latent encodes the agent’s interpretive stance. It is therefore orthogonal rather than competitive, and can coexist with existing decision stacks (Dreamer-like, LLM-based, or otherwise) as a layer for long-horizon coherence diagnostics, without replacing core control mechanisms.

\subsection{Possible Extensions to Language-Based Models}
\par The proposed architecture is not intended to compete with performance-optimized systems, but to coexist with them. Consistent with phenomenological accounts, subjectivity may shape behavior without being reducible to task performance \cite{Block1995, Zahavi2005-subjandselfhood}.

\par Because the contribution lies in structuring a slow latent state space, the same principle could extend to systems with internal representations, including LLM-based agents. A Perspective-like latent might track conversational regime shifts or long-horizon coherence, complementing token-level uncertainty measures.

\par Exploring such extensions remains future work. The present study clarifies what type of internal variable may be worth instrumenting when modeling perspective-like subjectivity.

\bibliography{aaai2026}
\end{document}